\documentclass[letterpaper, 10 pt, conference]{ieeeconf}  
\usepackage{hyperref}

\IEEEoverridecommandlockouts                              

\title{\LARGE \bf
A Portable Multiscopic Camera for Novel View and Time Synthesis in Dynamic Scenes
}

\author{Tianjia Zhang, Yuen-Fui Lau, and Qifeng Chen
\vspace{-0.6cm}
\thanks{Authors are with the Hong Kong University of Science and Technology, Hong Kong, China. T. Zhang (tzhangbl@connect.ust.hk) is with the Individualized Interdisciplinary Program and the Robotics Institute. Y. Lau (yflauad@connect.ust.hk) is with the Department of Mathematics. Prof. Q. Chen (cqf@ust.hk) is with the Department of Computer Science and Engineering and the Department of Electronic and Computer Engineering.}
}
\usepackage{graphicx}
\usepackage{cite}
\usepackage{amsmath}
\usepackage{lineno}
\usepackage[compact]{titlesec}
\begin{document}

\maketitle
\thispagestyle{empty}
\pagestyle{empty}

\vspace{-0.6cm}
\begin{abstract}
We present a portable multiscopic camera system with a dedicated model for novel view and time synthesis in dynamic scenes. Our goal is to render high-quality images for a dynamic scene from any viewpoint at any time using our portable multiscopic camera. To achieve such novel view and time synthesis, we develop a physical multiscopic camera equipped with five cameras to train a neural radiance field (NeRF) in both time and spatial domains for dynamic scenes. Our model maps a 6D coordinate (3D spatial position, 1D temporal coordinate, and 2D viewing direction) to view-dependent and time-varying emitted radiance and volume density. Volume rendering is applied to render a photo-realistic image at a specified camera pose and time. To improve the robustness of our physical camera, we propose a camera parameter optimization module and a temporal frame interpolation module to promote information propagation across time. We conduct experiments on both real-world and synthetic datasets to evaluate our system, and the results show that our approach outperforms alternative solutions qualitatively and quantitatively. Our code and dataset are available at \url{ https://yuenfuilau.github.io/}.

\end{abstract}

\section{INTRODUCTION}
Novel view synthesis (NVS) has been a trending research topic for 3D scene rendering and modeling, especially with the progress led by neural radiance fields (NeRF) \cite{mildenhall2020nerf}. NVS has potential real-life applications such as AR/VR in Metaverse, robotic vision system, aerial surveying, and autonomous driving \cite{chitta2021neat}. For instance, Block NeRF \cite{tancik2022block} has presented a 3D reconstruction approach to renders a 3D map on a city. Such an application can also be further extended to autonomous driving in terms of route planning and real-time environmental adaption. Currently, most existing approaches focus on novel view synthesis using multi-view images captured in a static scene or a monocular video of moving objects, which does not utilize multiscopic information over time. So we might want to ask: can we build a multiscopic camera tailored for novel view and time synthesis in dynamic scenes in the real world?


NVS in dynamic scenes is generally challenging, and most existing methods make some assumptions of captured input data. For instance, HyperNerf \cite{park2021hypernerf} assumes that the dynamics from a scene can be represented by a 2D latent vector, which limits the freedom of potential dynamic movements in a scene. Nerfies can only model a small amount of deformation of an object (i.e., head) \cite{park2021nerfies}.

In this work, we aim at solving novel view and time synthesis in a dynamic environment without any assumption about object movements, such as a room with walking persons, using a portable multiscopic camera that is accessible to general users. Our camera captures multi-view images at every time step (30 frames per second), which can provide abundant information for robust novel view and time synthesis. The portable multiscopic sensor is constructed with five cameras following the \textit{top-left-center-right-bottom} layout in the size of a laptop (about $30cm \times 30cm$) \cite{yuan2020mfusenet,yuan2021stereo}. The camera allows us to capture synchronized videos from five different perspectives. 

\begin{figure}
    \centering
    \includegraphics[width=0.51\textwidth]{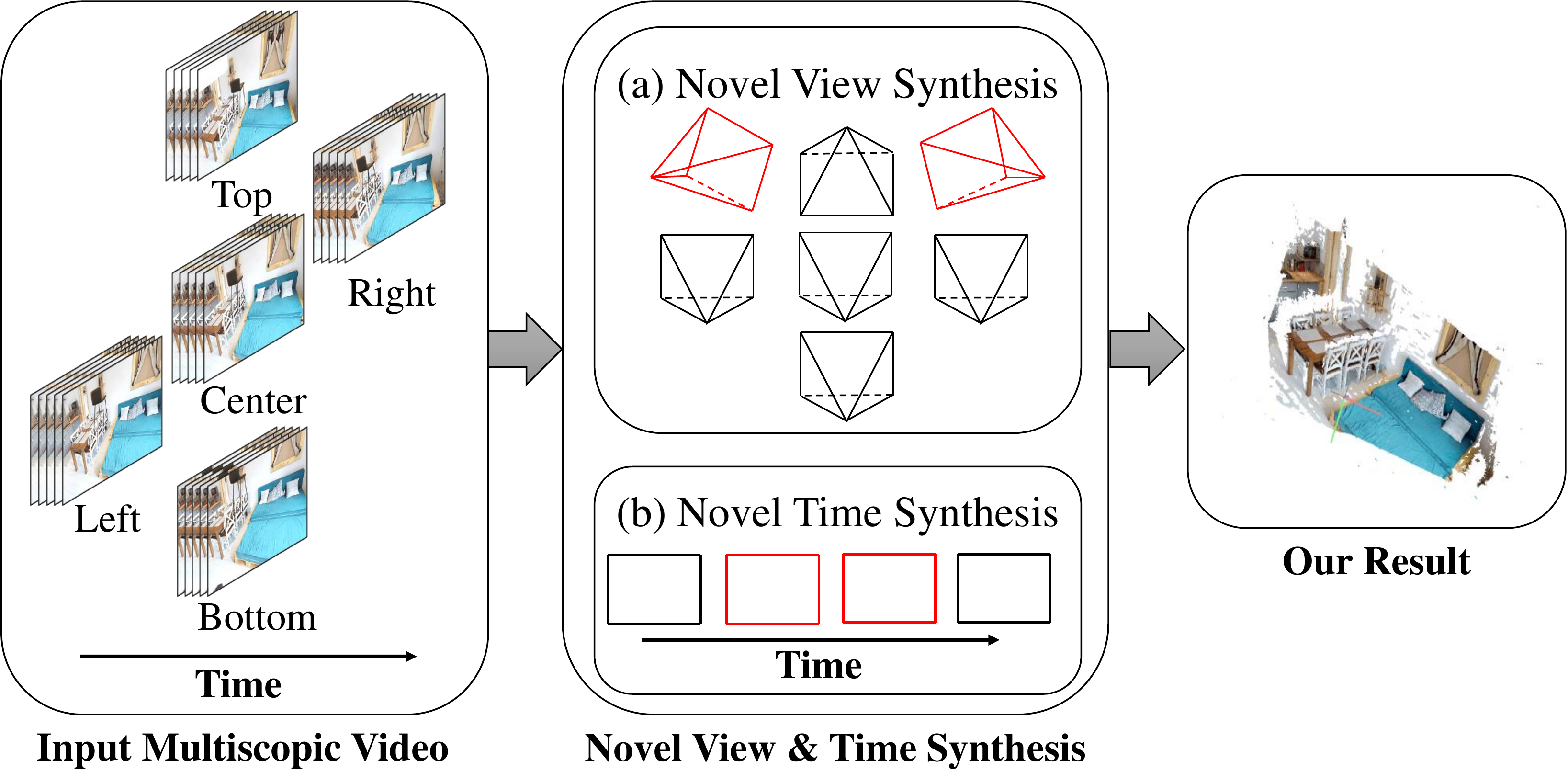}
    \caption{\textbf{Illustration of the objective of our multiscopic camera.} From several multiscopic input videos or images, we learn a time-varying scene representation to model the geometry and appearance of the scene. From such representation, we can generate novel images (red line) at any pose (a) and time (b). }
    \label{fig:taskl}
    \vspace{-0.6cm}
\end{figure}
\begin{figure}[t]
  \centering
  \includegraphics[scale=1.0]{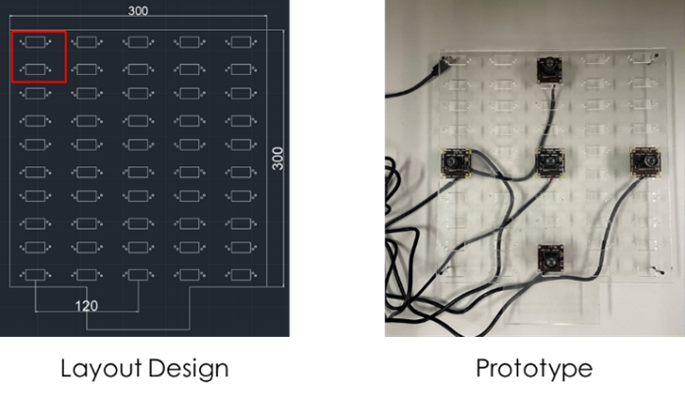}
    \vspace{-0.4cm}
  \caption{\textbf{The physical structure of our multiscopic camera.} The five cameras are fixed to the supporting board of size $30cm \times 30cm$, following the bottom-center-left-right-top layout.}
  \label{fig:camera_system}
  \vspace{-0.7cm}
\end{figure}

With our multiscopic camera, we propose an end-to-end NeRF-based method to perform novel view and time synthesis in dynamic scenes. Similar to previous time encoding work on NeRF \cite{pumarola2021d}, we extend NeRF to model emitted radiance and density of a point as a function of a 4D space-time coordinate and a 2D viewing direction. Compared to previous methods \cite{mildenhall2020nerf, du2020neuralflow, li2021neuralsceneflow}, which rely on either calibration or structure-from-motion method to obtain camera parameters, we embed the camera parameters and time-variable into a learnable model to be optimized during the training process, inspired by Wang et al. \cite{wang2021nerf}. To encourage information propagation across time,  we apply a video frame interpolation model SloMo \cite{jiang2018super} to generate several intermediate images between two consecutive frames and enforce our temporal NeRF model to render photo-realistic images at an arbitrary time. As a result, our model can perform not only novel view synthesis tasks but also novel time synthesis generated from our model. Fig. \ref{fig:taskl} illustrates the objective of our multiscopic camera, and Fig. \ref{fig:camera_system} shows the layout design and the physical prototype of our multiscopic camera.

To quantitatively evaluate our model for novel view and time synthesis, it is desirable to collect high-quality real-world images in a dynamic scene rendered from arbitrary viewpoints (beyond the five cameras) at any time, but such a data collection process requires lots of human effort and dedicated a large camera array system. Thus we build a synthetic dataset by applying a simulation engine to render images in a dynamic scene from any viewpoints. Our simulation system is based on the well-known Habitat-Sim system\footnote{Further details can be found at \url{https://github.com/facebookresearch/habitat-sim}.}, a modular intuitive and high-performance simulator for studying embodied AI. We can render high-quality images using 3D scene data with moving object (e.g., a flying chair) at a reasonable speed.

To summarize, our main contributions include
\begin{itemize}
    \item  A multiscopic camera prototype for novel view and time synthesis,
    \item A NeRF model trained end-to-end to perform novel view and time synthesis in dynamic scenes, at any camera pose and any time.
    \item A dataset of real-world data captured using the proposed camera and synthetic data rendered from customized simulators.
    \vspace{-0.4cm}
\end{itemize}
  
\section{RELATED WORK}
\noindent \textbf{Novel View Synthesis.}
Traditional methods on novel view synthesis are usually achieved by interpolation from densely distributed multiple perspectives. However, with the rise of deep learning, many researchers have applied various neural network structures to learn the synthesis process, and it has gradually become the mainstream research direction.
The key component to achieve successful view synthesis is to reconstruct an accurate and compact scene representation that could reflect the scene's geometry and visual appearance. 
At present, the most commonly used methods are volumetric representations \cite{seitz1999photorealistic} and multiple-plane images (MPI) \cite{zhou2018stereo,mildenhall2019local, srinivasan2019pushing}. 
Volumetric representation can be used to represent very complex scenes and objects. 
Researchers train neural networks to predict the color of voxels \cite{eric2017soft,kar2017learning} from images and then use rendering techniques \cite{porter1984compositing} to generate images. 
Additionally, MPI is a multi-layered scene representation consisting of a series of fronto-parallel planes at a fixed depth range in front of capturing the view. 
However, these methods are limited by discrete sampling. As the resolution increases, those methods cannot meet the needs of rendering ultra-high-definition images because the storage and loading time will increase sharply. \cite{ouyang2022real} have explored the application of MPI method on free-viewpoint avatar rendering of a moving person. However, it targets at dynamic rendering of dynamic characters and cannot adapt to arbitrary dynamic scenes.  The recent emergence of implicit representations\cite{niemeyer2019occupancy,mescheder2019occupancy} can model the scenes as a continuous function of spatial locations and other important geometry parameters. Those parameters can then be optimised via training.



\begin{figure*}[t]
\centering
\includegraphics[width=1\textwidth,height=0.4\textheight]{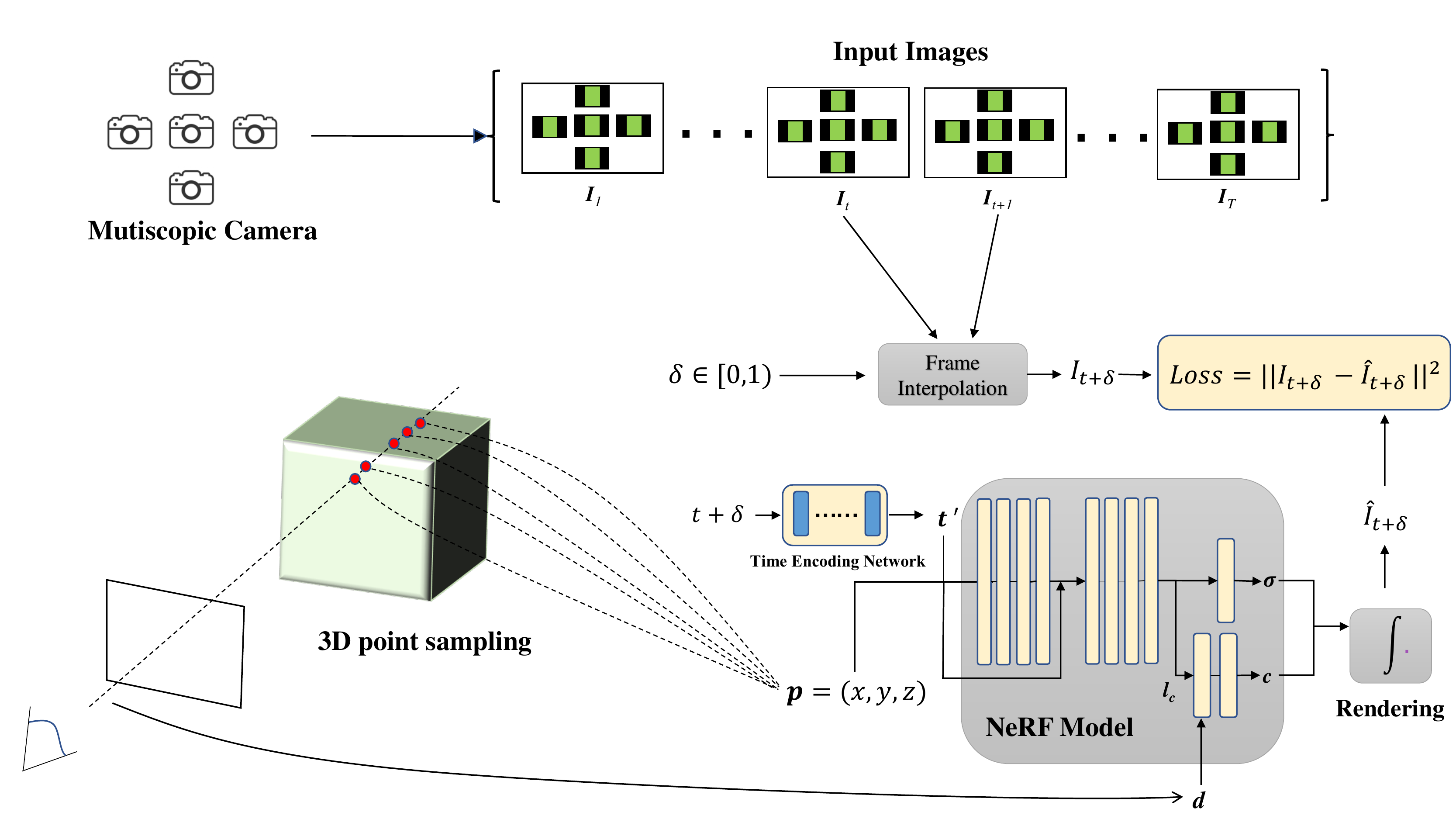}
\vspace{-0.8cm}
\caption{The pipeline of our method demonstrates the training and testing process, where $\delta \in $ [0,1). To render an image $\boldsymbol{\hat{I}_{t+\delta}}$ , given  optimized camera intrinsic and extrinsic parameters, an embedded latent time representation $\boldsymbol{t^{\prime}}$ obtained from the time encoding network and 3D points (\textbf{x},\textbf{y},\textbf{z}) sampled along camera rays are fed together into the NeRF Model with the viewing direction \textbf{d} to generate color radiance \textbf{c} and density $\boldsymbol{\sigma}$. The color \textbf{c} and density $\boldsymbol{\sigma}$ are then aggregated together to produce image $\boldsymbol{\hat{I}_{t+\delta}}$.}
\label{fig:pipeline}
\vspace{-0.6cm}
\end{figure*}

\noindent \textbf{Neural Radiance Field}
NeRF \cite{mildenhall2020nerf} is a neural implicit representation for modeling a single scene. It encodes
scenes and objects in the MLP so that continuous scenes can be represented at a relatively low cost.
However, due to the limit of the capacity of the MLP itself and the over-fitting optimization mechanism, NeRF can only be used to represent a single scene, which significantly limits its application. Many subsequent works were used to improve NeRF's representation ability and expand the scope of application in terms of generalization ability \cite{wang2021ibrnet}, dynamic and deformable scenes synthesis \cite{park2020deformable}, applications under extreme environments \cite{martin2021nerf}, improving training and rendering efficiency \cite{reiser2021kilonerf}, reconstruction on specific shape domains\cite{chen2021animatable, peng2021neural} and so on. Our interest focuses more on works extending NeRF to generate free-viewpoint videos.   
Inspired from level-set methods which use slices to represent the movement of surfaces, Hyper-NeRF\cite{park2021hypernerf} addresses the problem by lifting canonical NeRF into a higher dimensional space and slicing a cross-section to model topologically changing shapes. Several methods have been explored on extending NeRF to dynamic scenes. \cite{li2021neuralsceneflow, du2020neuralflow} combine flow models with NeRF to warp source frames to target frames and thus it could model the temporal dimension. \cite{xian2021space-time} shares a similar idea with us in that both represent the dynamic scenes as a function of space and time. 
However, these methods emphasize processing monocular videos since time-synchronized multi-view data is difficult to collect. 
Recently, cameras with image and depth sensors have been mass-manufactured due to the great demand for 3D reconstruction and autonomous driving. Additional information which helps capture scene geometry\cite{attal2021torf} could compensate the scale drawbacks of using visual information alone. However, depth information is not good at dense sensing and preserving visual consistency. Our method demonstrates that aggregating homogeneous visual information from multiple color sensors could achieve impressive reconstruction and rendering. 


\noindent \textbf{Multi-camera System and 3D Scene Simulator}
Humans evolve two eyes to better sense 3D geometry of the world. The biological phenomenon inspires the development of binocular stereo cameras. Nowadays stereo cameras are widely used in robotics and computer vision. However, two views could not provide sufficient constrain to estimate dense correspondence matching. Mathematicians\cite{hartley2003multiple} have been exploring abstract multi-view geometry for years, laying a solid theoretical foundation for the application of multi-view sensor system. Many works create camera arrays\cite{wilburn2005high} to enhance stereo matching between any two cameras and produce high-quality dense depth maps. However, those cumbersome camera systems are difficult to move. Other works\cite{yuan2020mfusenet, yuan2021stereo} use a single mobile camera to capture multiscopic data and apply the collected data to investigate depth estimation and stereo matching methods. 
Nonetheless, they cannot capture dynamic scene. \cite{broxton2020immersive} developed a spherical camera array consisting of 46 low-cost but wild-FOV mobile cameras. They designed extra software and hardware interface to synchronize internal timing. However, processing such a large amount of video data costs large memory and encodes redundant information. Our multiscopic camera system \cite{yuan2020mfusenet,yuan2021stereo} could balance portability, lightweight, and sufficient consistency constraints. 
Besides real-world data, synthetic data is also widely employed in researching synthesis methods. Some 3D datasets \cite{chang2017matterport3d,armeni20163d} and corresponding simulators \cite{puig2018virtualhome,wu2018building} have been proposed to visualize three-dimensional scenes, but their interactivity capabilities are inferior. Recently, with the emergence of simulators such as Gibson \cite{xia2018gibson} and Habitat-sim \cite{savva2019habitat,szot2021habitat} that focus on embodied AI tasks, it has become possible to train robots in simulators. Based on the powerful performance of the simulator, our method demonstrates that the simulators' data can be used for novel view and time synthesis which show great potential for computer vision tasks.
\section{Method}
Our goal is to render images or videos at arbitrary viewpoints and time given a set of multiscopic videos without knowing other prior information. Mathematically, We denote $\{(I_t^1, P_{t}^1), \hdots,(I_t^N, P_{t}^N)\}^{T}_{t=1}$ the source images and corresponding poses at time $t$, where $T$ and $N$ are the number of frames and the number of views ($N=5$), respectively. Our objective is to render a target image $\hat I_{t}$ with the specified target pose $P$ at time $t$. To address the problem, we formulate the neural radiance field as a function of a 4D space-time coordinate $(x,y,z,t)$ and a 2D viewing direction $\boldsymbol{d}$. The output consists of time-dependent radiance $c_t$ and volume density $\sigma_t$. To exploit and integrate the visual consistency information across time, we apply the frame interpolation method\cite{jiang2018super} to generate intermediate frames, which serve as extra ground-truth to supervise the training process. If the camera parameters are unknown, we simultaneously optimize the intrinsic parameter in normal space and the extrinsic parameters in SE(3) space. 


\subsection{NeRF Model} \label{subsectionNerf}
NeRF \cite{mildenhall2020nerf}  applies an MLP network to predict color radiance and density for an input 5D  vector consisting of a 3D position vector and a 2D viewing direction vector. The color radiance represents the RGB value at the location, and the density represents the probability that a camera ray terminates at that location. The viewing direction is used to cope with viewing-dependent effects such as reflection and diffusion. However, this amount of information is insufficient for NeRF to render a correct image in a dynamic scene. The time dependence is essential for the sake of novel view synthesis via time. Hence, the input for our NeRF is slightly different from the standard one. Since it deals with a changing environment over time, an extra time coordinate $t$ will be introduced to allow our model to represent time domain. Additionally, taking inspiration from a parameterized pose learning network \cite{wang2021nerf}, we apply a latent time representation.The time coordinate $t$ is warped using an MLP to get a higher level representation, allowing the model to distinguish between frames at each moment more easily. The latent time representation is given by
\vspace{-0.2cm}
\begin{equation}
	\begin{aligned}
		\boldsymbol{t^{\prime}} = \mathcal{W}(t), 
		\label{MLP}
	\end{aligned}
 \vspace{-0.2cm}
\end{equation}

\noindent where $\mathcal{W}$ is a time encoding network,
and $\boldsymbol{t^\prime}$ is a latent representation of the time coordinate.

Since the density of a point solely relies on the space and time coordinate, the model divides the mapping relation into two parts: predicting density using only the space-time coordinate and predicting radiance color using both the space-time coordinate and viewing direction. Mathematically, the relation could be represented by a two-stage MLP:
\vspace{-0.1cm}
\begin{equation}
	\begin{aligned}
		\sigma, \boldsymbol{l_c} &= \text{MLP}_1(\boldsymbol{p},\boldsymbol{t}^{\prime}), \\
		\boldsymbol{c} &= \text{MLP}_2(\boldsymbol{l_c}, \boldsymbol{d}),
	\end{aligned}
\label{MLP_model}
\vspace{-0.1cm}
\end{equation}

\noindent where $\boldsymbol{p}=(x,y,z)$ is the 3D spatial coordinates; $\text{MLP}_1$ comprises a 8-layer network which transforms the 4D input into the scalar density $\sigma$ and a 256-D hidden feature vector $\boldsymbol{l_c}$; the feature vector $\boldsymbol{l_c}$ is concatenated with the unit viewing direction $\boldsymbol{d}$ and then fed into the final layer $\text{MLP}_2$; the radiance color $\boldsymbol{c}$ is the output in terms of RGB value. 
Following NeRF \cite{mildenhall2020nerf}, volume rendering is applied to produce the pixel color of a ray passing through the scene. Specifically, for one pixel, a camera ray is cast originated at the camera center. For each point in the ray we can predict its density and radiance using the model Eq. \ref{MLP_model}. The color of a pixel is obtained by integrating the weighted product of the radiance value and the density along the camera ray. However, it is difficult to computer the continuous integral. Fortunately, quadrature is a powerful tool to approximate the integral. Suppose a camera ray $\boldsymbol{r}$ has origin $\boldsymbol{o}$ and unit viewing direction $\boldsymbol{d}$, we sample $\mathcal{Z}$ points 
on the ray $\boldsymbol{r}(z) = \boldsymbol{o}+z\boldsymbol{d}$ within the near and far scene bound $z_n$, $z_f$.  The estimated color along the ray $\boldsymbol{r}$ is calculated by 
\vspace{-0.2cm}
\begin{equation}
\hat{C}(\boldsymbol{r}) = \sum^{\mathcal{Z}}_{i=1}T_{i}(1-e^{-\sigma_i\Delta_i})\boldsymbol{c}_{i},
\vspace{-0.2cm}
\end{equation}
\noindent where $T_{i}=\exp(-\sum^{i-1}_{j=1}\sigma_j\Delta_j )$ and $\Delta_i = z_{i+1} - z_{i}$ is the distance between $z_{i+1}$ and $z_{i}$.

Our method synthesizes images at given poses and time steps. We do not render a full image during training since it would run out of memory. Instead, we randomly cast camera rays for corresponding image pixels using the optimised camera parameters and render RGB values with our model. In general, our losses are evaluated in the sample space using mean square error:
\vspace{-0.3cm}
\begin{equation}
    \mathcal{L}_{MSE} = \sum_t^T\sum_i^N\| I_{t}^i - \hat{I}_{t}^i  \|^2,
    \vspace{-0.2cm}
\end{equation}

\noindent where $I_{t}$ is the sampled ground-truth image at step $t$ and $\hat{I}_{t}$ is the rendered image outcome.



\subsection{Frame Interpolation} \label{subsectionflow}

It is intuitive to directly add the time variable to the input and transform the 4D space-time coordinates using the MLP network. However, we found that each dimension would entangle with the other. Consequently, the simple mapping alone cannot model such a complex relationship and results in image degradation through time. We disentangle the time and space by using an optical-flow-based SloMo model to enforce pixel consistency in the temporal domain.

For the $i$-th sensor (out of $N$) in our multiscopic camera, given two adjacent frames $I_t$ and $I_{t+1}$ (we regard $I_t$ as $I_t^i$ for simplicity in the following description), we apply the Super SloMo method\cite{jiang2018super} to render an intermediate frame ${I_{t+\delta}}$  ($0\leq \delta<1$) and use the 4D-input NeRF model (Eq. \ref{MLP_model}) to predict a frame using the same settings, we encourage the two generated frame to be similar to enhance temporal connection. 


 We exploit neural networks to predict the optical flow between the intermediate target and source frames. The frame is generated by interpolating the images warped from initial images using the flow field. However, the target frame image is not known initially, which indicates we cannot obtain the optical flow field from the source frames to the target frame. To deal with the issue, we assume that the motion of the dynamic scenes is slow and smooth. Therefore the target flow fields could be estimated from the weighted average of the bidirectional flow fields $F_{t \rightarrow t+1}$ and $F_{t+1\rightarrow t}$.

The estimation works poorly near the motion boundaries and cannot address the occlusion issue. As FlowNet2 \cite{ilg2017flownet} indicates, the cascaded CNN architecture is beneficial to refine the predicted results and reduce artifacts. In addition, the visibility maps, which represent whether a pixel is visible when moving from one step to another, are used to handle occlusions. Therefore, the refining process takes the coarse optical flow prediction and source color images as input and outputs the refined optical flow fields and visibility maps.  

Since the input and output of the flow estimation and refinement process are all grid maps, the mapping relation follows the encoder-decoder architecture and can be modeled using a
U-Net \cite{ronneberger2015unet} convolutional neural network. 
In particular, given $\hat{F}_{t+\delta \rightarrow t+1}$ and $\hat{F}_{t+\delta \rightarrow t}$ that are coarse predicted flow fields and $g(I_{t+1}, \hat{F}_{t+\delta \rightarrow t+1})$ and $g(I_t, \hat{F}_{t+\delta \rightarrow t}))$ that represent synthesizing potential target images by warping from source images $I_{t}$ and $ I_{t+1}$ using the corresponding coarse predicted flow fields, respectively, we then feed them into another U-Net to generate the predicted flow fields, visibility maps $V$, and refined RGB images. Finally, those flow features output are aggregated together as demonstrated in Super SloMo approach\cite{jiang2018super} to produce the intermediate frame:
\vspace{-0.3cm}
\begin{equation}
\begin{aligned}
I_{t+\delta}=&\frac{1}{\lambda}\odot[(1-\delta)V_{t\rightarrow t+\delta}\odot g(I_t,\hat{F}_{t+\delta \rightarrow t}) +\\
& \delta V_{t+1 \rightarrow t+\delta}\odot g(I_{t+1},\hat{F}_{t+\delta \rightarrow t+1})].
\end{aligned}
\end{equation}
where $\lambda = (1-\delta)V_{t+\delta \rightarrow t} + \delta V_{t+\delta \rightarrow t+1}$ is normalization factor and $\odot$ denotes the element-wise multiplication.


In addition, the temporal consistency loss is added to enforces our model which is learnt from the flow models inherently to find the correspondence of pixels across time. We randomly select two consecutive frames from a single input video. We use our model to render the pixels sampled from the intermediate frame given the estimated pose $P$ and time step $t$. At the same time, the flow model could interpolate between the selected frames and output the warped RGB values for the same pixels. The operations used in our model are all differentiable and can be performed end-to-end. Fig. \ref{fig:pipeline} illustrates how to train our model from scratch and produce photo-realistic novel images from the trained model. The temporal consistency loss is the mean squared error of the two values, which is calculated by
\begin{equation}
\begin{aligned}
I_{t+\delta} &= \textit{Interpolation}(I_t, I_{t+1}, \delta), \\
\mathcal{L}_{t+\delta} &= \|\hat I_{t+\delta}-I_{t+\delta}\|^2.
\end{aligned}
\end{equation}

\subsection{Optimizing Camera Parameters}

Due to vibration and assembly tolerance, the camera parameters would not stay constant across time in real-world cases. Following NeRF-- \cite{wang2021nerf}, we jointly optimize the network model and the camera parameters. The extrinsic matrix locates in the $SE(3)$ space and is not closed for addition operation. We optimize the parameters in $se(3)$ space and transform them back to $SE(3)$, so that the parameters are valid during optimization. An vector in $se(3)$ consists of a 3D translation vecotr $\boldsymbol{n}$ and a 3D rotation vector $\boldsymbol{r} = [r_x, r_y, r_z]$. The transformation from a rotation vector to a rotation matrix is the \textit{Rodrigues' Rotation Formula}, which is described by
\begin{equation} \label{Rformula}
    \boldsymbol{R} = \boldsymbol{I} + sin(\theta)\boldsymbol{K} + (1-cos(\theta))\boldsymbol{K^2} ,
\end{equation}
where $\boldsymbol{I}$ is the identity matrix, $\theta$ is the length of the rotation vector, and $\boldsymbol{K}$ is the skew-symmetric matrix derived from $\boldsymbol{r}$.






\section{Experiments}
\subsection{Experiment Setting}
\noindent\textbf{Multiscopic Camera System}
Pilot work\cite{yuan2020mfusenet, yuan2021stereo} uses a single camera mounted on a moving robotic arm to simulate a multi-scopic camera. Although this setting can guarantee identical intrinsic parameters for different views and proper extrinsic parameters, its application is limited to in-door scenarios. Based on the previous work, we built a portable 5-view real-world camera to enable casual and dynamic capturing. We manufactured an acrylic support to mount five SONY IMX 322 camera modules with almost the same configurations. We place one module on the \textit{center} position and place other modules equally distant from the center module in the four directions: \textit{top}, \textit{bottom}, \textit{left} and \textit{right}. At present, we set the distance between adjacent modules 120mm. As we have reserved much space on the supporting board, the positions and number of modules could be adjusted according to practical needs.

\noindent\textbf{Synthetic Data Generator} Our customized real-world camera could capture five views at one time. However, we require more views to evaluate the performance of our NVS method quantitatively. The quality of real-world data suffers from device noise and improper exposure. Moreover, the depth information is usually difficult to estimate without dedicated sensors. To address the issue, we developed a simulated multi-view (no longer limited to 5 views) camera based on the well-known Habitat-sim simulator designed for embodied robotic tasks. The current works use the simulator mainly for testing robots' performance in autonomous navigation and instruction following. Our work provides a more impressive and exciting application scenario for the simulator.
To enable dynamic simulation, we keep the agent and the cameras still and place a chair inside the common viewing frustum of the cameras. The chair would move from the left side to the right side from the capturing perspective. 

\noindent\textbf{Data Preparation} We collected a set of real-world and synthetic data from various scenes using the sensor and the simulator mentioned above. For synthetic data, we render 24 frames of data per scene. Each frame contains five images rendered using the 5-view layout for training 
For real-world data, we captured 120 frames of 5-view data per scene. 
The image resolutions are $1280\times720$ for synthetic data and $640\times480$ for real data, respectively.

\noindent\textbf{Environment and Parameters} Our experiments are conducted on a server equipped with Intel Xeon 5218 CPU and Nvidia RTX 2080 Ti GPU. We implement our network model in Pytorch and the data reading interface in Python. We use the Adam optimizer to train the MLP network and the camera parameters for 1200 epochs. We fix the flow model and load the pretrained checkpoint \footnote{The implemention of the flow-based frame interpolation model is available at:
https://github.com/avinashpaliwal/Super-SloMo} since we want to learn the temporal consistency embedded in the model. For any two consecutive frames, 4 intermediate frames are generated from the frame interpolation model for temporal consistency training. The learning rate is set at 0.001 initially and exponentially decays every 100 epochs unless reaching $1e-5$. We sample 6400 rays and 128 points along each ray during the training to render pixel colors for each iteration.

\subsection{Results}
We conduct extensive experiments to demonstrate the ability of our method to achieve photo-realistic novel view and time synthesis. More importantly, we compare our approach to state-of-the-art flow- and NeRF-based dynamic novel view synthesis methods\cite{li2021neuralsceneflow,du2020neuralflow} to show that our proposed approach achieves comparable performance.
\begin{figure*}[t]

\centering
\begin{tabular}{c@{\hspace{0.5mm}}c@{\hspace{0.5mm}}c@{\hspace{0.5mm}}c@{\hspace{0.5mm}} c@{}}
\includegraphics[width=0.245\linewidth,height=0.16\linewidth]{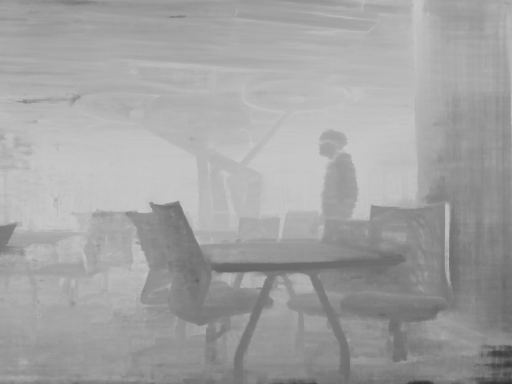}&
\includegraphics[width=0.245\linewidth,height=0.16\linewidth]{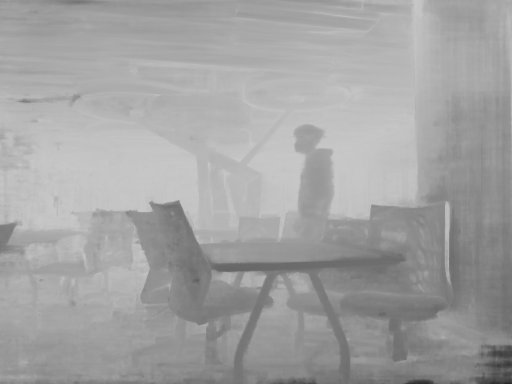}&
\includegraphics[width=0.245\linewidth,height=0.16\linewidth]{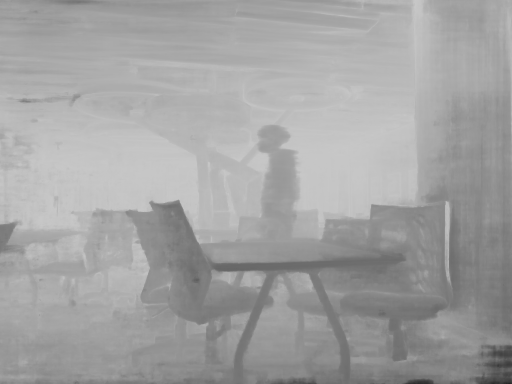}&
\includegraphics[width=0.245\linewidth,height=0.16\linewidth]{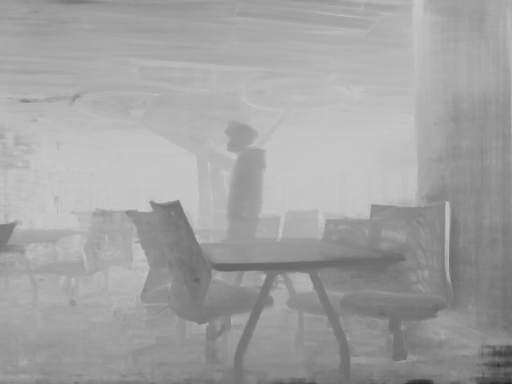}
\\
\includegraphics[width=0.245\linewidth,height=0.16\linewidth]{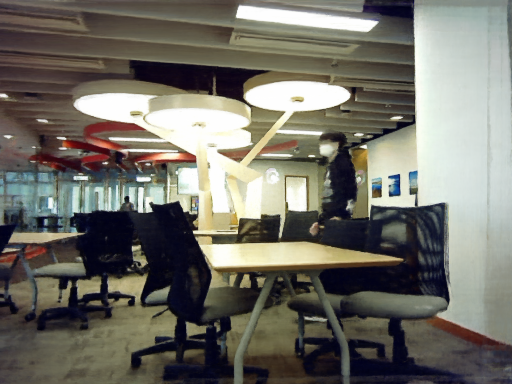}&
\includegraphics[width=0.245\linewidth,height=0.16\linewidth]{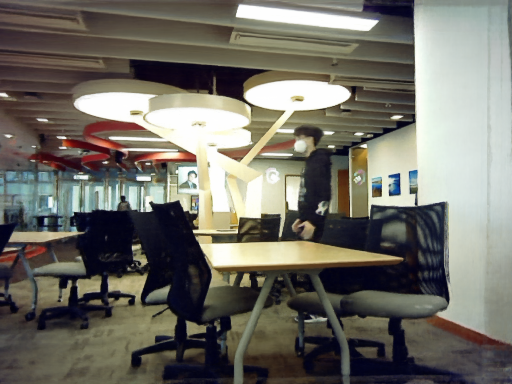}&
\includegraphics[width=0.245\linewidth,height=0.16\linewidth]{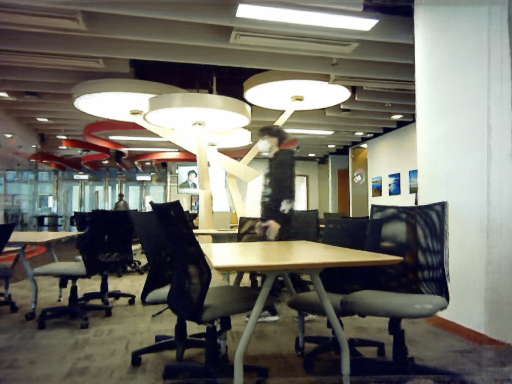}&
\includegraphics[width=0.245\linewidth,height=0.16\linewidth]{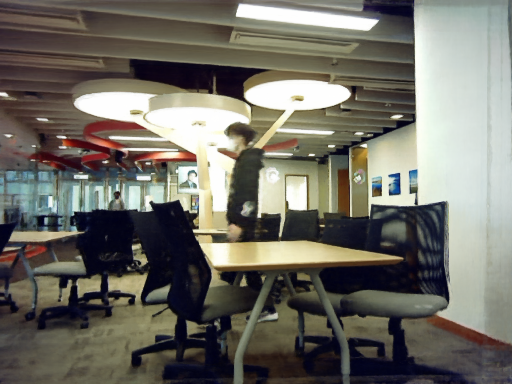}
\\
Time 1 & Time 2 & Time 3 & Time 4 
\\
\includegraphics[width=0.245\linewidth,height=0.144\linewidth]{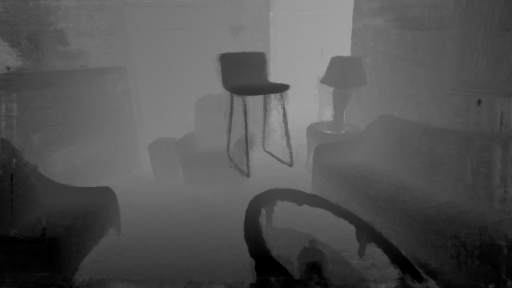}&
\includegraphics[width=0.245\linewidth,height=0.144\linewidth]{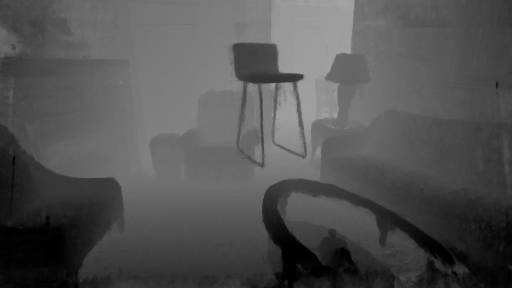}&
\includegraphics[width=0.245\linewidth,height=0.144\linewidth]{images/room2_depth1_SlMo.png}&
\includegraphics[width=0.245\linewidth,height=0.144\linewidth]{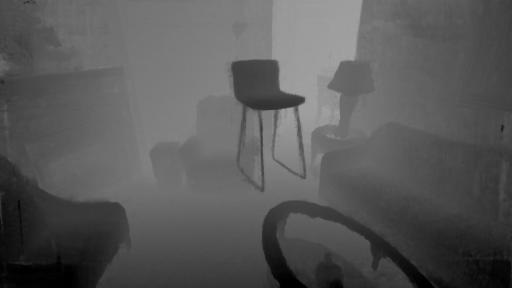}
\\
\includegraphics[width=0.245\linewidth,height=0.144\linewidth]{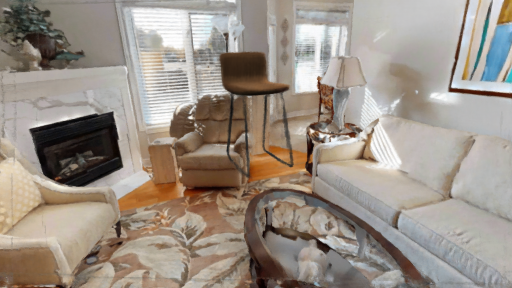}&
\includegraphics[width=0.245\linewidth,height=0.144\linewidth]{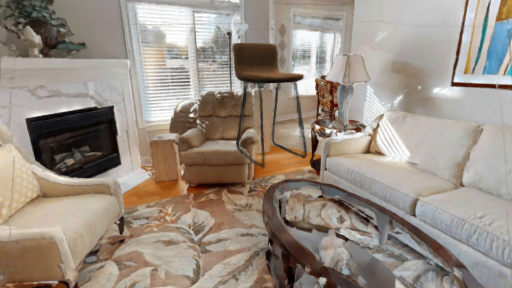}&
\includegraphics[width=0.245\linewidth,height=0.144\linewidth]{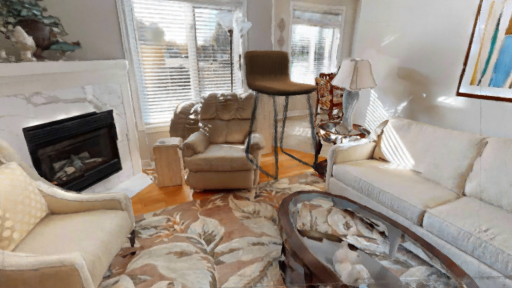}&
\includegraphics[width=0.245\linewidth,height=0.144\linewidth]{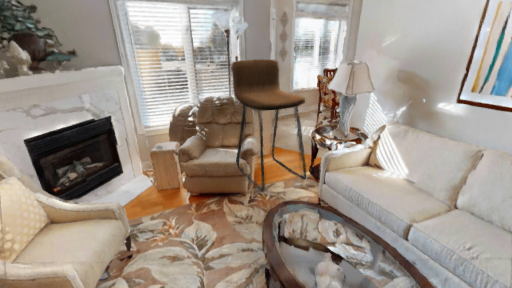}
\\
View 1 & View 2 & View 3 & View 4 
\end{tabular}
\vspace{-0.4cm}
\caption{\textbf{Rendering Result} with variation in time at fixed camera pose and variation in camera pose at a fixed time.}
\label{fig:novel_view_time_syn}
\vspace{-0.7em}
\end{figure*}

\begin{figure}[t]
\centering
\begin{tabular}{c@{\hspace{0.5mm}}c@{\hspace{0.5mm}}}
\includegraphics[width=0.45\linewidth,height=0.23\linewidth]{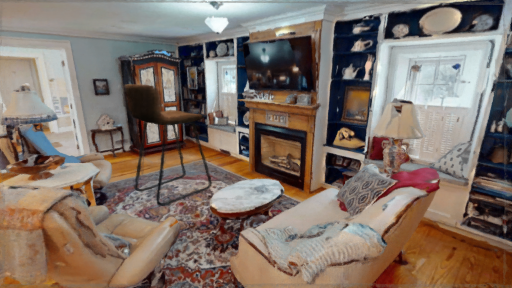}&
\includegraphics[width=0.45\linewidth,height=0.23\linewidth]{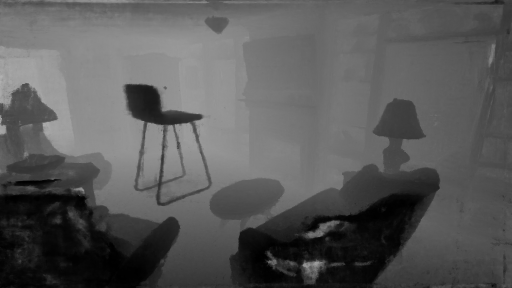}
\\
\includegraphics[width=0.45\linewidth,height=0.23\linewidth]{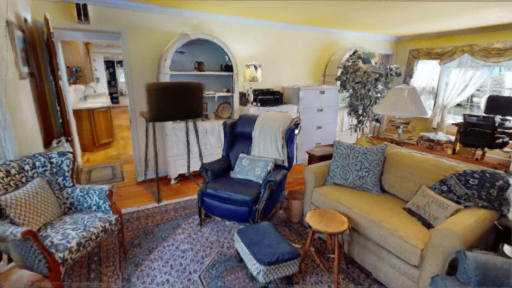}&
\includegraphics[width=0.45\linewidth,height=0.23\linewidth]{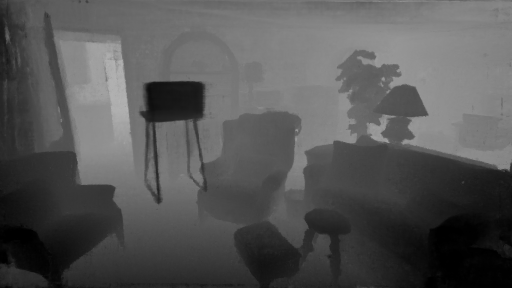}
\\
RGB Image & Depth Map\\
\end{tabular}
\vspace{0mm}
\caption{\textbf{More results of depth maps rendered using our method from complex scenes}. We have no access to any ground-truth depth map, and thus, the estimation is completely performed in a self-supervised fashion.}
\label{fig:depth_map}
\vspace{-0.8cm}
\end{figure}

\noindent\textbf{Illustration of Novel View and Time Synthesis} We present rendered images at unseen viewpoints and specified timesteps in Fig. \ref{fig:novel_view_time_syn}.  We synthesize four images for the fixed time and camera pose cases, respectively. The results illustrate our method's generalization ability in space and time domains. It proves that most fine-grained appearance details are well embedded in our space-time model. Moreover, we leverage the predicted density to render depth maps to show that our model can capture adequate geometry information. More results to visualize depth quality in complex scenes are listed in Fig. \ref{fig:depth_map}. Since we have no ground-truth depth supervising signal during the training process, our method could serve as an auxiliary tool for self-supervised depth estimation or multi-view stereo matching. 
\\
\begin{figure*}[t]
\centering
\begin{tabular}{c@{\hspace{0.5mm}}c@{\hspace{0.5mm}}c@{\hspace{0.5mm}}c@{\hspace{0.5mm}}}
\includegraphics[width=0.245\linewidth,height=0.144\linewidth]{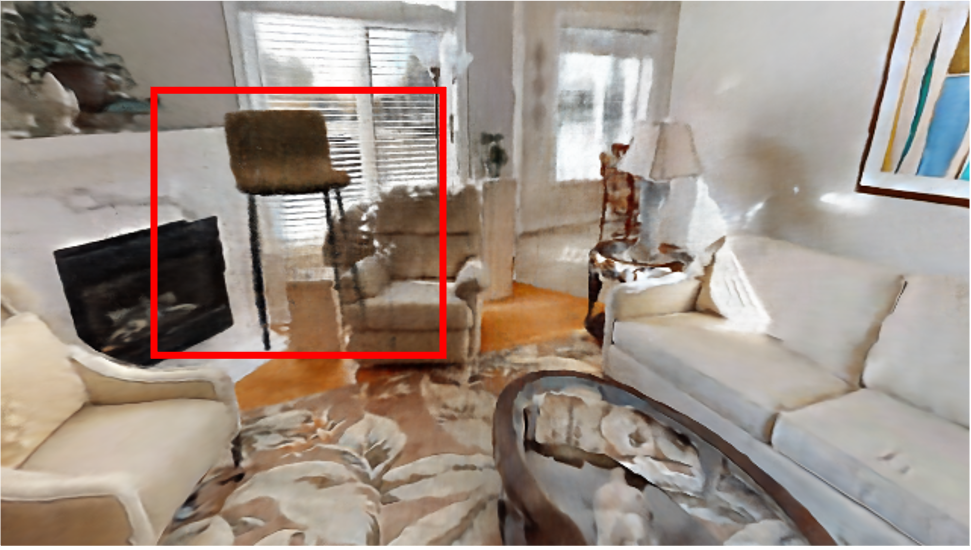}&
\includegraphics[width=0.245\linewidth,height=0.144\linewidth]{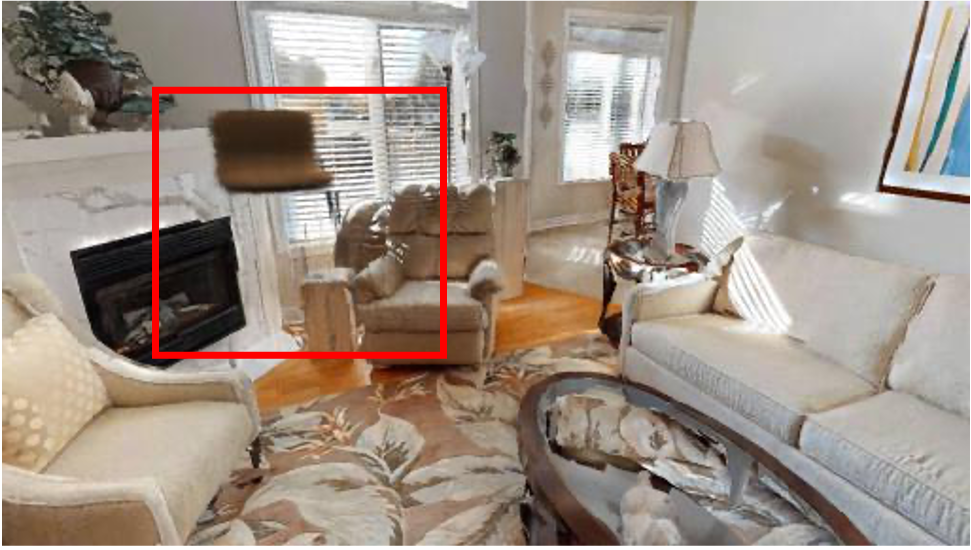}&
\includegraphics[width=0.245\linewidth,height=0.144\linewidth]{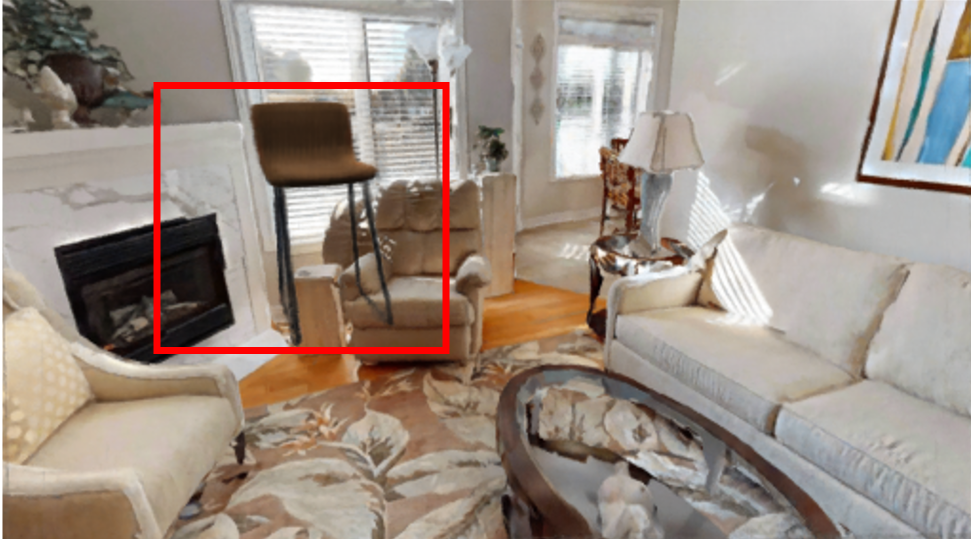}&
\includegraphics[width=0.245\linewidth,height=0.144\linewidth]{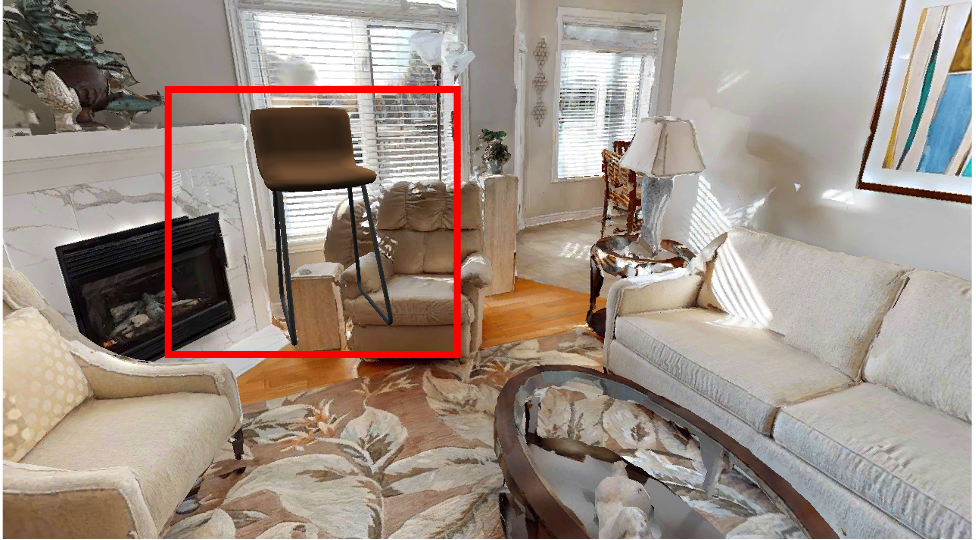}
\\
\includegraphics[width=0.245\linewidth,height=0.144\linewidth]{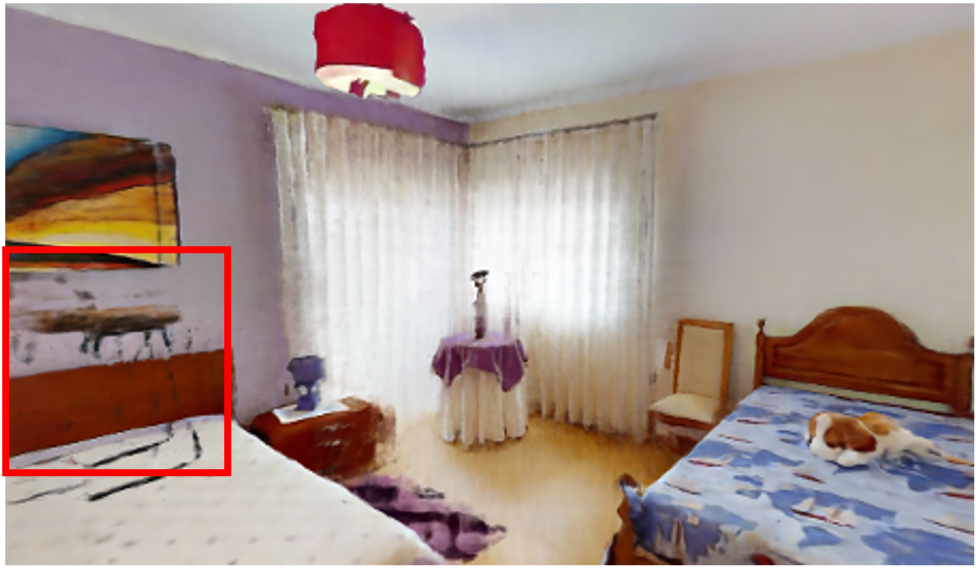}&
\includegraphics[width=0.245\linewidth,height=0.144\linewidth]{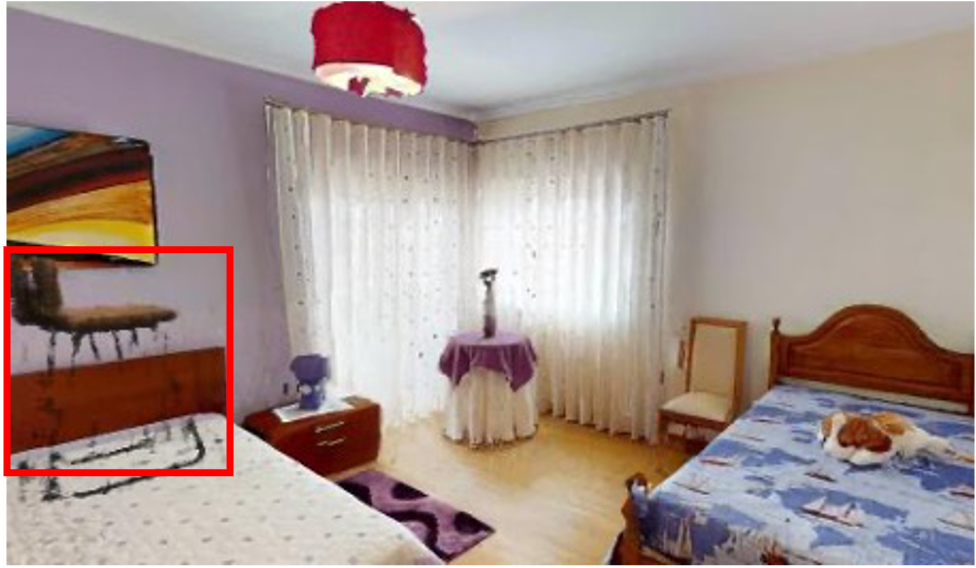}&
\includegraphics[width=0.245\linewidth,height=0.144\linewidth]{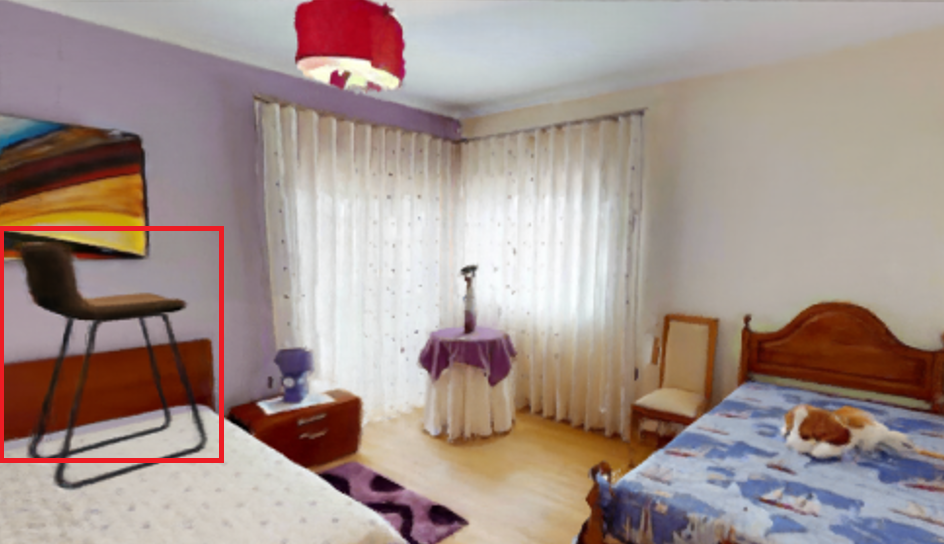}&
\includegraphics[width=0.245\linewidth,height=0.144\linewidth]{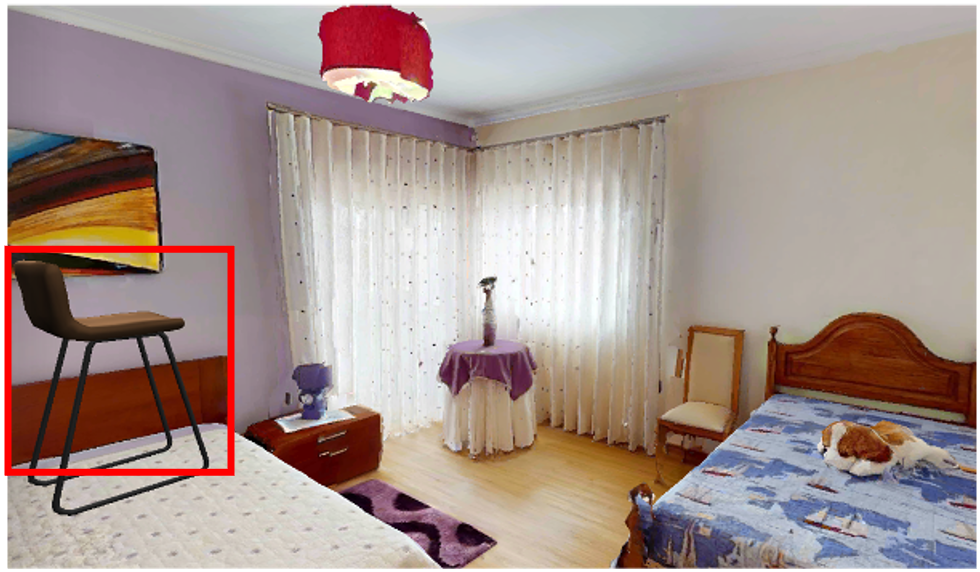}
\\
NeRFlow & NSFF & Ours & Ground Truth \\
\end{tabular}
\vspace{-0.3cm}
\caption{\textbf{Qualitative comparisons on synthetic data.}. The dynamic scenes are generated following the same motion pattern: a chair moves from the left to the right. The reconstructed images are rendered from a randomized pose at a randomized time step. We compare our method with NeRFlow\cite{du2020neuralflow} and NSFF\cite{li2021neuralsceneflow}.All methods are able to render high-quality static background scenes, while our method could reconstruct the moving chair better with less blurry artifacts and reconstruction failures.}
\label{fig:quali_syn}
\vspace{-0.7em}
\end{figure*}
\begin{figure*}[t]
\centering
\begin{tabular}{c@{\hspace{0.5mm}}c@{\hspace{0.5mm}}c@{\hspace{0.5mm}}c@{\hspace{0.5mm}} c@{}}
\includegraphics[width=0.245\linewidth,height=0.16\linewidth]{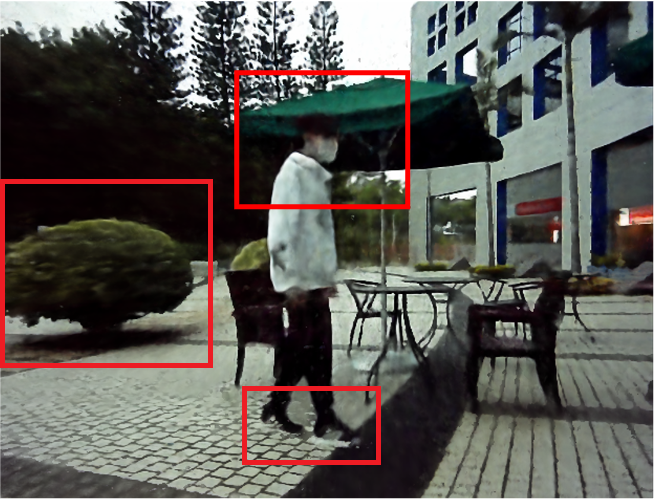}&
\includegraphics[width=0.245\linewidth,height=0.16\linewidth]{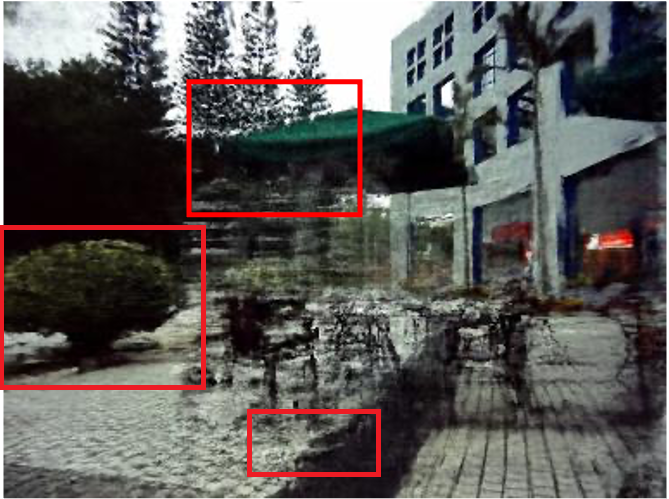}&
\includegraphics[width=0.245\linewidth,height=0.16\linewidth]{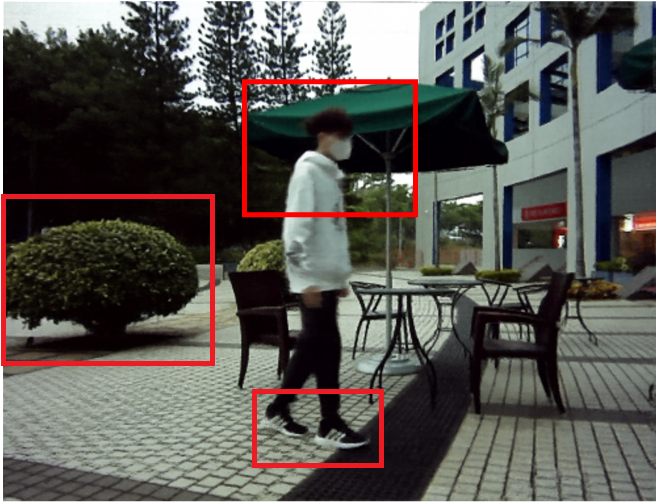}&
\includegraphics[width=0.245\linewidth,height=0.16\linewidth]{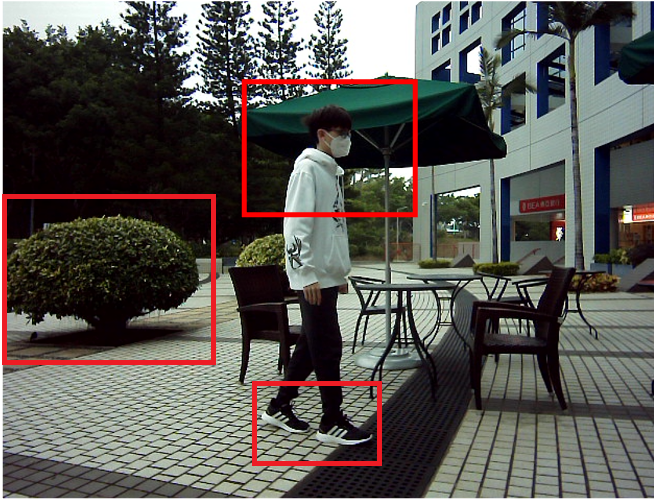}
\\
\includegraphics[width=0.245\linewidth,height=0.16\linewidth]{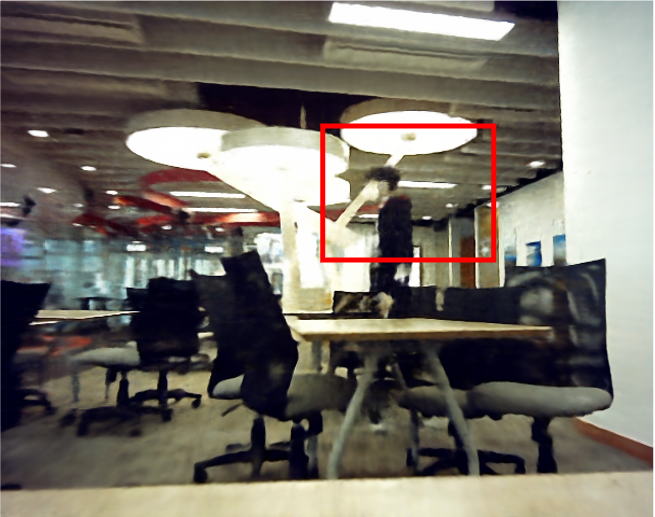}&
\includegraphics[width=0.245\linewidth,height=0.16\linewidth]{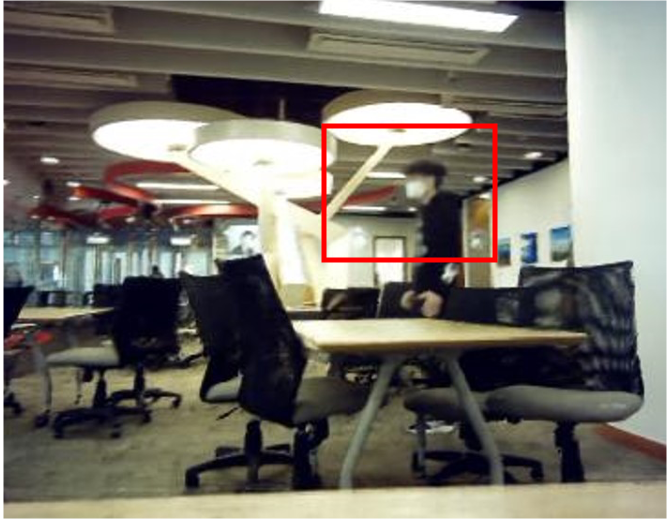}&
\includegraphics[width=0.245\linewidth,height=0.16\linewidth]{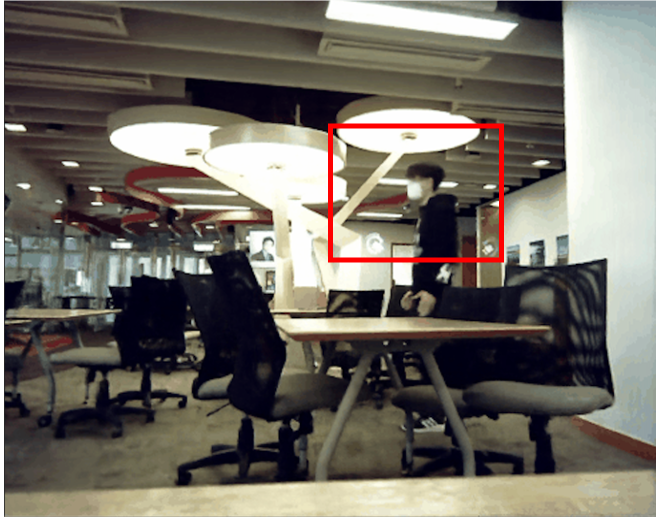}&
\includegraphics[width=0.245\linewidth,height=0.16\linewidth]{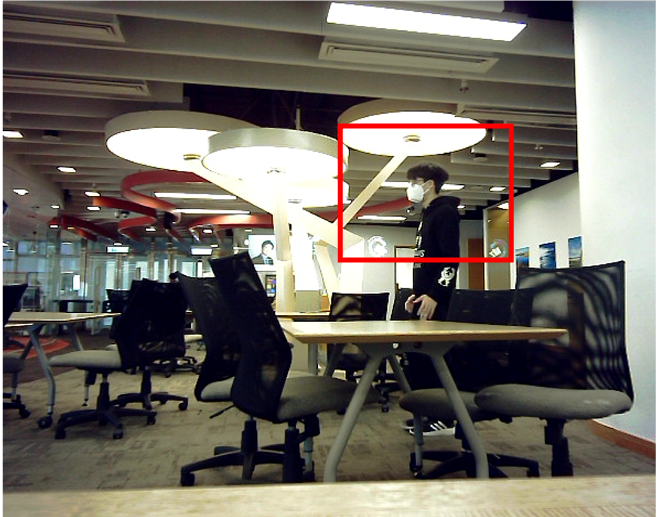}
\\
NeRFlow & NSFF & Ours & Ground Truth\\
\end{tabular}
\vspace{-0.4cm}
\caption{\textbf{Qualitative comparisons on real data}. To capture the data, we fix the multi-scopic camera sensor and make a person move in front of the background scenes. We render the novel images following the same settings of synthetic scenarios. The grass area and the human head area are marked by red rectangles to show that our method can render finer details. }
\label{fig:quali_real}
\vspace{-0.7em}
\end{figure*}

\noindent\textbf{Qualitative and Quantitative Comparison}
We carefully select top-performing methods NSFF\cite{li2021neuralsceneflow} and NeRFlow\cite{du2020neuralflow} as the baseline to evaluate our model. The ideas of their methods are similar to ours in that both represent the dynamic scene as a function of space-time coordinates. Nevertheless, all the methods propose a different mechanism to enforce coherent temporal learning. We train their models from scratch on our data, following the instructions from the officially published repositories. All parameters are set as default.   
For quantitative evaluation, we report two error metrics describing the similarity between the rendered images and the ground-truth images: structural similarity index measure (SSIM) and peak signal-to-noise ratio (PSNR). The metric statistics are shown in Table \ref{tab:quantitativesyn},\ref{tab:quantitativereal}, which demonstrates that our method outperforms the baseline methods in all metrics. 


\renewcommand{\arraystretch}{1.2}
\begin{table}[]
    \centering
    \caption{Quantitative evaluation on synthetic data}
    \begin{tabular}{c c c c}
    \hline
    Method   & PSNR $\uparrow$ & SSIM $\uparrow$    \\
    \hline
    NeRFlow \cite{du2020neuralflow}   & 27.32 & 0.75\\
    NSFF \cite{li2021neuralsceneflow}     &27.31 & 0.73 \\
    Ours     &\textbf{33.06} & \textbf{0.88}  \\
    \hline
    \end{tabular}
    \label{tab:quantitativesyn}
    \vspace{-0.2cm}
\end{table}

\renewcommand{\arraystretch}{1.2}
\begin{table}[]
    \centering
    \caption{Quantitative evaluation on real data}
    \begin{tabular}{c c c c}
    \hline
    Method   & PSNR $\uparrow$ & SSIM $\uparrow$  \\
    \hline
    NeRFlow \cite{du2020neuralflow}   & 27.61 & 0.66   \\
    NSFF \cite{li2021neuralsceneflow}     &26.39 & 0.64\\
    Ours     &\textbf{29.68} & \textbf{0.75}   \\
    \hline
    \end{tabular}
    \label{tab:quantitativereal}
    \vspace{-0.8cm}
\end{table}

We perform qualitative analysis on both synthetic and real datasets to find where our method takes effect. Fig. \ref{fig:quali_syn} and Fig.\ref{fig:quali_real}  illustrate the rendering results. The synthetic results show that all the methods can reconstruct the static environment background, which demonstrates the powerful capability of NeRF model on static novel view synthesis. However, NSFF is bad at rendering small or thin foreground objects, e.g., the support of the chair. NeRFlow preserves the global content better while bringing more noisy points. Our rendered images avoid such artifacts and become more photo-realistic. For real-world cases, we compare performance in typical indoor and outdoor scenes. NeRFlow and NSFF show similar results as those of synthetic cases. It should be noted that both baseline methods have failure cases. NSFF fails at outdoor scenes and generates messy and chaotic foreground objects, while NeRFlow fails at scene boundary. We attribute the artifacts to the improper estimation of camera poses and motion fields during training, highlighting the necessity of our two proposed refining modules.


\renewcommand{\arraystretch}{1.2}
\begin{table}[]
    \centering
    \caption{Ablation study on synthetic data}
    \begin{tabular}{c c c c}
    \hline
    Method   & PSNR $\uparrow$ & SSIM $\uparrow$   \\
    \hline
    NeRF (without time) &24.13 & 0.78  \\
    Ours without optimization     &29.25 & 0.74  \\
    Ours without $\mathcal{L}_t$    &32.43 & 0.70  \\
    Ours     &\textbf{33.06} & \textbf{0.88}  \\
    \hline
    \end{tabular}
    \label{tab:quantitativesyn}
    \vspace{-0.2cm}
\end{table}
\renewcommand{\arraystretch}{1.2}
\begin{table}[]
    \centering
    \caption{Ablation study on real data}
    \begin{tabular}{c c c c}
    \hline
    Method   & PSNR $\uparrow$ & SSIM $\uparrow$   \\
    \hline
    NeRF (without time) &21.51 & 0.63  \\
    Ours without optimization     &28.58 & 0.74  \\
    Ours without $\mathcal{L}_t$    &26.17 & 0.59  \\
    Ours     &\textbf{29.68} & \textbf{0.75}   \\
    \hline
    \end{tabular}

    \label{tab:quantitativereal}
    \vspace{-0.7cm}
\end{table}

\noindent\textbf{Ablation Study}
We evaluate the impact of the proposed modules individually in the task of dynamic novel view synthesis by removing (1) latent time representation (NeRF (without time)); (2) optimizing camera parameters (without optim); (3) the temporal consistency loss  (without $\mathcal{L}_t$). The results, as given in Table \ref{tab:quantitativesyn} and \ref{tab:quantitativereal}, reveal the relative significance of each component, with the entire system doing the best. We observed that each component of our approach contributed to a different improvement in the outcome. We found that without the reinforcement of temporal consistency, image deterioration becomes considerable, resulting in a large drop in SSIM value and a slightly decrease in perceptual similarity. The decline in SSIM value and perceptual similarity can be attributed to the model's failure to disentangle space temporal information without the aid of flow model. On the other hand, we discovered that time encoding is essential in obtaining a decent PSNR result since the model might fail to query frames accurately. Without appropriate latent time representation, NeRF model tends to ignore  dynamic part of the scene and focus on optimizing the static part only. This causes additional noise and inconsistency in the dynamic scene, resulting in a low PSNR. Furthermore, because the fraction of dynamic objects is tiny in comparison to the whole picture, the SSIM value drops little owing to the degradation of the dynamic component.

\section{Conclusion}

In this work, we build a portable multiscopic camera sensor for novel view and time synthesis using an end-to-end NeRF-based model that takes synchronized input views from the multiscopic camera. To evaluate our model, we built a synthetic dataset as well as a real-world dataset with dynamic scenes rendered by a multiscopic camera. Our NeRF-based model learns a mapping from a space and time representation to radiance and density for rendering a dynamic scene. We conduct an extensive evaluation on both synthetic and real-world datasets and the results show that our model outperforms alternative solutions by a large margin. With our customized NeRF-based model, we hope our new multiscopic camera designed for novel view and time synthesis can be a practical module in robot applications.

\addtolength{\textheight}{4cm}   





\bibliographystyle{IEEEtran.bst}
\bibliography{IEEEfull.bib}








\end{document}